\documentclass[aps,12pt,nofootinbib]{revtex4-2}
\usepackage{amsmath}
\usepackage{amsthm}
\usepackage{mathtools}
\usepackage{hyperref}
\usepackage{amssymb}
\usepackage{physics}
\usepackage{graphicx}
\graphicspath{ {images/} }
\usepackage[caption=false]{subfig}
\usepackage{float}

\usepackage{enumitem}
\usepackage{comment}

\usepackage{tikz}
\usetikzlibrary{fit}
\usetikzlibrary{trees}
\usetikzlibrary{positioning}

\usepackage[section]{placeins} 

\DeclareMathOperator*{\argmax}{argmax}

\newcommand{\etal}{\textit{et al}. }

\DeclareMathOperator{\ch}{ch} 
\DeclareMathOperator{\gch}{gch} 
\DeclareMathOperator{\p}{p} 
\DeclareMathOperator{\s}{s} 

\newcommand{\beq}{\begin{equation}}
	\newcommand{\eeq}{\end{equation}}

\begin{document}

	\title{An efficient solution to Hidden Markov Models on trees with coupled branches}
	\author{Farzan Vafa}
	\affiliation{Center of Mathematical Sciences and Applications, Harvard University, Cambridge, USA}
        \affiliation{Department of Data Science, Dana-Farber Cancer Institute, Boston, MA, USA}
	\author{Sahand Hormoz}
        \email{sahand_hormoz@hms.harvard.edu}
	\affiliation{Department of Systems Biology, Harvard Medical School, Boston, MA, USA}
	\affiliation{Department of Data Science, Dana-Farber Cancer Institute, Boston, MA, USA}
        \affiliation{Broad Institute of MIT and Harvard, Cambridge, MA, USA}
        
	\begin{abstract}
	Hidden Markov Models (HMMs) are powerful tools for modeling sequential data, where the underlying states evolve in a stochastic manner and are only indirectly observable. Traditional HMM approaches are well-established for linear sequences, and have been extended to other structures such as trees. In this paper, we extend the framework of HMMs on trees to address scenarios where the tree-like structure of the data includes coupled branches— a common feature in biological systems where entities within the same lineage exhibit dependent characteristics. We develop a dynamic programming algorithm that efficiently solves the likelihood, decoding, and parameter learning problems for tree-based HMMs with coupled branches. Our approach scales polynomially with the number of states and nodes, making it computationally feasible for a wide range of applications and does not suffer from the underflow problem. We demonstrate our algorithm by applying it to simulated data and propose self-consistency checks for validating the assumptions of the model used for inference. This work not only advances the theoretical understanding of HMMs on trees but also provides a practical tool for analyzing complex biological data where dependencies between branches cannot be ignored.
	\end{abstract}
	\maketitle
	
	
	\section{Introduction}
	
    Hidden Markov Models (HMMs) are routinely used for statistical inference of sequences of states where states evolve in a stochastic manner and can only be observed indirectly. The parameters of these models are the probability of the initial state being a particular hidden state, the probability of transition from one hidden state to another at each time step, and the probability of an observation conditional on a given hidden state. In most applications, the sequence of the hidden states takes the form of a linear chain.
    
    Efficient algorithms have been proposed to compute the likelihood of a set of observations given the model parameters (likelihood problem), the most likely set of hidden states given the model parameters (decoding problem), or inferring the model parameters from sequences of observations (learning). In the case of a chain of states, the likelihood, decoding, and learning problems can be solved by the forward-backward~\cite{rabiner1989tutorial}, Viterbi~\cite{viterbi1967error}, and Baum-Welch algorithms~\cite{baum1966statistical,baum1967inequality,baum1970maximization}. These algorithms use dynamic programming to compute the answer efficiently. For example, naively, we might expect that to compute the likelihood of a set of observations, we must sum over all possible hidden states. If there are $N$ possible hidden states and a sequence of $T$ states, we would need to sum over $T^N$ states. With dynamic programming, the summation over hidden states is written as a recursion and computed instead with only $\mathcal O(TN^2)$ operations.

    Hidden Markov Models have been extended beyond simple linear chains and applied to other structures, such as hierarchical models~\cite{fine1998the}, coupled models~\cite{brand1997coupled}, factorial models~\cite{ghahramani1995factorial}, and more generally graphical models~\cite{koller2009probabilistic}. We will focus on the extension of Hidden Markov Models to trees introduced by Crouse \etal~\cite{crouse1998wavelet}. The specific application that they had in mind was to capture the hierarchical interdependence of coefficients of wavelet transforms. In tree HMMs, each state corresponds to a node of the tree. As with conventional HMMs, the state of each node can only be indirectly observed. The states of the descendants of a node, $h_i$, on the tree are conditional on the state of their parent node, $h_0$, and given by the transition matrix, $P(h_i|h_0)$. Importantly, the state of each descendant is assigned independently from the states assigned to the other descendants; the branches of the tree are uncoupled. That is, in the case of two descendant nodes, $P(h_1,h_2|h_0) = P(h_1|h_0)P(h_2|h_0)$.

    Here, we extend the concept of a tree HMM to the case where the branches of the tree connecting a parent node to its children are coupled, namely, $P(h_1,h_2|h_0) \neq P(h_1|h_0)P(h_2|h_0)$. At first, this might seem like an odd choice; a tree by definition contains branches that are independent from each other. Why consider coupled branches? Our motivation stems from trees encountered routinely in biology where branches connecting sister nodes are not independent. Consider for example dividing cells. One cell divides into two cells, each of which then divides to two more cells, and so on, generating a binary tree. Each cell is in a molecular state that determines its phenotype, for example its morphology or the duration of its cell cycle. The molecular states of the daughter cells clearly depend only on the molecular state of their mother cell, justifying a Markovian model~\cite{hormoz2016inferring,hughes2022patterns,Mohammadi2022}. However, the molecular state of the two sisters cells can be coupled even if conditioned on the state of their parent. For example, if one daughter cell receives too many copies of a molecule present in the mother cell then the other daughter cell will receive fewer copies. Models of trees with independent branches will not capture such correlations~\cite{Hormoz2014}. Therefore, we need to solve hidden Markov models on tree with coupled branches.

    A significant problem when applying dynamic programming to HMMs is the so called underflow problem. The underflow problem arises during the computation of probabilities over long chains of sequences, which includes multiplying transition probabilities and observation probabilities together repeatedly. For long sequences, the repeated multiplication of probabilities can result in small numbers that are beyond the precision range of floating-point representation in computers, resulting in numerical instabilities. To address this issue, Devijver~\cite{Devijver85} proposed an innovative solution by scaling the intermediate probabilities at each step of the forward and backward pass along the chain. By scaling these probabilities by a factor that keeps the sum of the probabilities across all hidden states at each time-step equal to one, the probabilities are prevented from becoming too small and causing underflow. This algorithm was extended by Durand \etal~\cite{durand2004computational} to HMMs on trees. We extend the results of Durand \etal~\cite{durand2004computational} to trees with coupled branches.
    
    The contributions of this paper are as follows. 1) We present an efficient solution using dynamic programming to the problem of hidden Markov models on trees with coupled branches. Surprisingly, the coupling of the branches does not preclude a solution with polynomial time in the size of the tree. We show that the computational complexity only increases for binary trees with $T$ nodes and $N$ hidden states from $\mathcal O(TN^2)$ to $\mathcal O(TN^3)$ when branches are coupled. Our results are general and can be applied to trees that have nodes with arbitrary number of descendants and trees where the number of descendants vary across the nodes. 2) We extend the results of Durand \etal~\cite{durand2004computational} to trees with coupled branches providing an efficient solution that does not suffer from the underflow problem. 3) Finally, we present an implementation of our algorithm in Python and apply it to simulated data as validation.

	\section{Model}
	\label{sec:model}
	
	\subsection{Definitions / elements of model}

 Hidden Markov models on trees require three key variables: the tree structure representing the familial connections of nodes (assumed to be an outward directed rooted tree), observations, and hidden states. In the case of dividing cells, a binary tree represents the cell's familial connections, the observations can include data like cell division time and size, while the hidden states are variables we cannot directly measure, such as chemical concentrations.
	
    We need to prescribe how these variables interact with each other. The transition probability $P(h_1, h_2 | h_0)$ describes the joint probability distribution of the hidden states $h_1$ and $h_2$ of the children, given the hidden state $h_0$ of the parent. The form of $P(h_1, h_2 | h_0)$ embodies the Markov property, namely, that the joint probability distribution of the hidden states of a node and its sibling's depends only on the hidden state of their parent (not for instance of their grandparent, cousins, or grandchildren.).
	
	The emission probability $P(O | h)$ represents the probability of observing a particular property $O$, given the current hidden state $h$ of the node. The form of $P(O | h)$ assumes output independence, i.e., that the probability of an observation at a node depends solely on its current hidden state, and not on any observations or any other hidden states.

\begin{figure}[htbp]
\centering
\includegraphics[width=\linewidth]{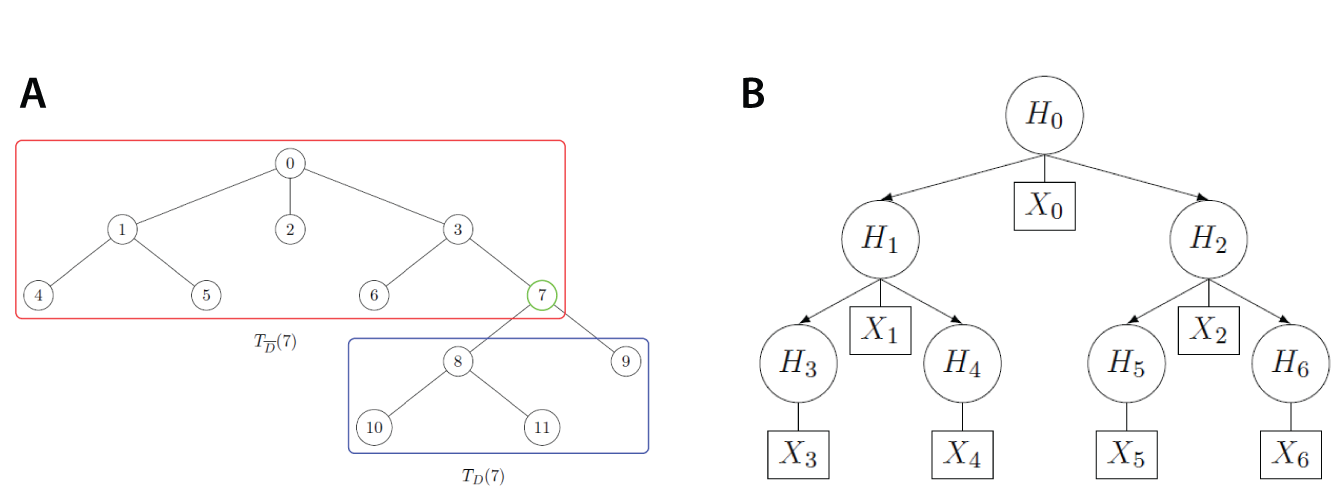}
\caption{Diagram of a rooted tree $T$ where 0 denotes the root. A. The leaves $T_L = \{2, 4, 5, 6, 9, 10, 11\}$ and the interior $T_I = \{0, 1, 3, 7, 8\}$. As an example for node $7$, the subtree rooted at node $7$ is $T_R(7) = \{7, 8, 9, 10, 11\}$, its parent $\p(7) = 3$, its children $\ch(7) = \{8,9\}$, its sibling $\s(7) = 6$, and its grandchildren $\gch(C) = \{10, 11\}$. Also, its descendants $T_D(7)$ is denoted by the blue box and the complement $T_{\overline D}(7)$ by the red box. B. Schematic of HMT, with nodes labeled 0 to 6, where circles represent hidden states and squares represent observations. For each node $C$, the observation $O(C) = X_C$ and the hidden state $h(C) = H_C$.}
\label{fig:schematic}
\end{figure}

	We now formally define the tree geometry and establish terminology. In this paper, we consider outward directed rooted trees, i.e., all the edges point away from the root. Familiar examples include chains and binary trees. Let $T$ denote such a tree, which has $|T|$ nodes. We introduce the following notation for a tree $T$, described pictorially in Fig.~\ref{fig:schematic}A:
	\begin{itemize}
		\item $0$ is the root of $T$.
		\item $T_L$ are the leaves of $T$, i.e., the nodes that have no children.
		\item $T_I \equiv T\setminus T_L$ is the interior of $T$ (the entire tree except for the leaves). 
	\end{itemize}
	
	We also introduce the following notation for a given node $C$, again described pictorially in Fig.~\ref{fig:schematic}A:
	\begin{itemize}
		\item $\p(C)$ is the parent of node $C$.
		\item $\ch(C)$ is the children of node $C$.
		\item $\s(C)$ is the siblings of node $C$.
		\item $\gch(C)$ are the grandchildren of node $C$.
		\item $T_D(C)$ is the descendants of node $C$, i.e. the maximal subtree of $T$ with $C$ as a root and excluding $C$ itself. Explicitly, $T_D(C) \equiv \{\ch(C), \gch(C),\ldots\}$.
        \item $T_R(C)$ is the maximal subtree of $T$ rooted at $C$, i.e., $T_R(C) \equiv \{C, T_D(C)\} = \{C, \ch(C), \gch(C),\ldots\}$.
		\item $T_{\overline D}(C)$ is the complement of $T_D(C)$, i.e., $T_{\overline D}(C) \equiv T\setminus T_D(C)$. $T_{\overline D}(C)$ can also be interpreted as the maximal subtree of $T$ with $C$ as a leaf.
  		\item $T_{\overline R}(C)$ is the complement of $T_R(C)$, i.e., $T_{\overline R}(C) \equiv T\setminus T_R(C)$.
	\end{itemize}

	We now define the variables on $T$. For notational convenience, we will denote hidden states by Greek letters and observations by Latin letters. We use the general notation $P()$ to denote a probability density function. A Hidden Markov Tree Model (HMT) is characterized by the following:
	
	\begin{enumerate}
		\item An outward directed rooted tree $T$, which has $|T|$ nodes.
		\item $N$, the number of hidden states in the model. We denote the individual states as $H = \{H_1, H_2, \ldots, H_N\}$, and the state at node $C$ as $h(C)$.
		\item We denote the observation at node $C$ as $O(C)$.
		\item For a node $C$ which has $n$ children, the state transition probability distribution $a^{\mu_0}_{\mu_1\ldots \mu_n}$, where 
		\beq a^{\mu_0}_{\mu_1\ldots \mu_n} = P(h(\ch(C)_1) = \mu_1, h(\ch(C)_2) = \mu_2, \ldots, h(\ch(C)_n) = \mu_n | h(C) = \mu_0), \eeq
		with the following probability constraints:
		\begin{subequations}
			\begin{align}
				a^{\mu_0}_{\mu_1\ldots \mu_n} &\ge 0 \\
				\sum_{\mu_1\mu_2\ldots\mu_n} a^{\mu_0}_{\mu_1\ldots \mu_n} &= 1 ,
			\end{align}
		\end{subequations}
		where $1 \le \mu_0, \mu_1, \ldots, \mu_n \le N$.
        
        The key difference between our formulation of HMTs compared with existing models is that the probability of the state of the children conditional on the state of their parent is coupled (or equivalently, the branches emanating from a parent node to its children are coupled). That is, in general,
        \beq a^{\mu_0}_{\mu_1\ldots \mu_n} \neq \prod_{i=1}^n P(h(\ch(C)_i) = \mu_i| h(C) = \mu_0). \eeq
		\item For a node $C$, the observation probability distribution given state $\mu$, $b_\mu(v)$, where
		\beq b_\mu(v) = P(O(C) = v | h(C) = \mu) .\eeq
		\item The initial state distribution $\pi(\mu)$ of the root, where
		\beq \pi(\mu) = P(h(0) = \mu), \qquad 1 \le \mu \le N .\eeq
	\end{enumerate}
	
	Thus to define a HMT, we need the number of hidden states $N$, as well as the three probability measures, $a$, $b$, and $\pi$, which we specify compactly by $\lambda$ as $\lambda = (a, b, \pi)$. See Fig.~\ref{fig:schematic}B for a diagram illustrating some of these definitions.

	\subsection{Three Fundamental Problems for HMTs}
	
	In general, there are three types of problems that we would like to solve for HMTs:
	\begin{enumerate}[wide=1em,  leftmargin=13em]
		\item[\bf{Problem 1 (likelihood)}:] Given the observations $O = O(T)$ and a model $\lambda = (a, b, \pi)$, efficiently compute the likelihood $P(O|\lambda)$.
		\item[\bf{Problem 2 (decoding):\,\,}] Given the observations $O = O(T)$ and a model $\lambda = (a, b, \pi)$, ``optimally'' determine the hidden states $h(T)$.
		\item[\bf{Problem 3 (learning): \,}] Given the observations $O = O(T)$, efficiently learn the model parameters $\lambda = (a, b, \pi)$ to maximize $P(O|\lambda)$.
	\end{enumerate}
	
	Problem 1 in the HMM literature is known as the likelihood problem~\cite{rabiner1989tutorial}, i.e. given a
	model and a tree of observations, how do we compute the probability that the model produces the observed tree, or how do we compute the likelihood of the observed tree? We can view the solution to this problem as how well our model predicts an observed tree, which allows us to choose the model among several competing models that best predicts the observed tree.
	
	Problem 2 in the HMM literature is known as the decoding problem~\cite{viterbi1967error}, i.e., given a
	model and a tree of observations, how do we find the ``correct'' tree of hidden states? Generally, there is no single ``correct'' tree of hidden states. Hence we will suggest and use an optimality criterion to best solve this problem. As in the case of HMM, there are several reasonable optimality criteria that we can impose, and therefore the intended use will determine the optimality criteria.

	Problem 3 in the HMM literature is known as the learning problem~\cite{baum1970maximization}, i.e., given a model and a tree of observations, how do we optimize the
	model parameters to best describe the observations? The observed trees can be seen as training data used to ``train'' the HMT. This problem is crucial since it allows us to optimally adapt model parameters to observed data, i.e., to create the best models for observed phenomena.
	
	In the next section we present solutions to each of the three fundamental problems. We will see that they are tightly linked.
	
	\section{Solutions to the three fundamental problems of HMTs}
    \label{sec:algorithms}
    
\subsection{Solution to Problem 1}

We wish to compute the probability of the observed tree $O=O(T)$ given the model $\lambda$, i.e. the likelihood $P(O|\lambda)$. The brute-force method is to enumerate over every possible tree $h$ of hidden states:
\beq P(O|\lambda) = \sum_h P(O, h|\lambda) = \sum_h P(O|h,\lambda)P(h|\lambda) .\eeq
We first note that
\beq P(O|h,\lambda) = \prod_{C \in T} P(O(C) | h(C), \lambda) ,\eeq 
where we have explicitly assumed that the observations are independent and depend only on the associated hidden state. We also note that $P(h | \lambda)$, the probability of tree of hidden states, can be written as
\beq P(h|\lambda) = \pi({h(0)})\prod_{C \in T_I} a^{h(C)}_{h(\ch(C)_1) \ldots h(\ch(C)_{|\ch(C)|})} .\eeq
Thus we have
\beq P(O|\lambda) = \sum_h \pi({h(0)}) 
\left(\prod_{C' \in T} P(O(C') | h(C')\right)\left(\prod_{C \in T_I} a^{h(C)}_{h(\ch(C)_1) \ldots h(\ch(C)_{|\ch(C)|})}\right) .\label{eq:likelihood-brute-force}\eeq

By inspection, Eq.~\eqref{eq:likelihood-brute-force} requires $\mathcal O(N^{|T|})$ computations, which even for small values of $N$ and $|T|$ quickly becomes unfeasible. Clearly, a more efficient procedure is needed to solve Problem 1. We now present such a solution, which is an extension of the forward-backward procedure~\cite{rabiner1989tutorial} and the ``upward-downward" algorithm introduced in Ref.~\cite{crouse1998wavelet} but generalized to trees with coupled branches.

Similar to HMMs, we will define the two variables: the backward probability, and the forward probability. However, unlike HMMs, the recursive definition of the forward probability depends on the backward probability. Hence we will first define the backward probability, and then define the forward probability in terms of the backward probability. For simplicity, we assume trees where the number of children for each node (other than the leaves) is fixed to be $n$. In the case of a binary tree, $n=2$. Our results can be easily generalized to the case where the number of descendants vary across the nodes of the tree.

We first define the backward probability
\beq \tilde\beta_C(\rho) \equiv P(O(T_R(C))|h(c) = \rho,\lambda) \eeq
to be the probability of observing $O(T_R(C))$, the maximal observed subtree of $T$ with $C$ as a root, given that node $C$ is in hidden state $\rho$ and the model $\lambda$. $\tilde\beta_C(\rho)$ can be expressed recursively as
\beq \tilde\beta_C(\rho) = b_{\rho}(O(C))\sum_{\mu_1 \ldots \mu_n}  a^\rho_{\mu_1\ldots\mu_n}\prod_{i=1}^n\tilde\beta_{\ch(C)_i}(\mu_i) , \qquad C \in T_I , \label{eq:beta}\eeq
where the termination condition is that $\tilde\beta_L(\rho) = b_{\rho}(O(L))$ on a leaf $L$ of $T$. By inspection, Eq.~\eqref{eq:beta} requires $\mathcal O(|T| N^{n+1})$ computations. These computations are done recursively. $\tilde\beta_C(\rho)$ at the leaves are directly obtained from the emission functions. Then we move up the tree and perform the summation in Eq.~\eqref{eq:beta} to obtain the $\tilde\beta_C(\rho)$ of the parent nodes. This is iterated until we reach the root of the tree.

We also define the forward probability 
\beq \tilde\alpha_C(\rho) \equiv P(O(T_{\overline R}(C)), h(C) = \rho|\lambda) \eeq
to be the probability of node $C$ being in hidden state $\rho$ after observing $O(T_{\overline R}(C))$, the complement of $T_R(C)$, i.e., $T_{\overline R}(C) \equiv T \setminus T_R(C)$. $\tilde\alpha_C(\rho)$ can be expressed recursively as
\beq \tilde\alpha_C(\rho) = \sum_{\mu_0\ldots\mu_{n-1}} b_{\mu_0}(O(\p(C))\tilde\alpha_{\p(C)}(\mu_0) a^{\mu_0}_{\rho \mu_1 \ldots \mu_{n-1}} \prod_{i=1}^{n-1}\tilde\beta_{\s(C)_i}(\mu_i). \label{eq:alpha}\eeq
Denoting the root of the tree by $0$, initially we have
\beq \tilde\alpha_0(\rho) = \pi(\rho) ,\eeq 
where  $\pi(\rho)$ is the initial hidden state distribution. By inspection, again Eq.~\eqref{eq:alpha} requires $\mathcal O(|T| N^{n+1})$ computations. This time the computations are done iteratively starting at the root of the tree and moving down to the child nodes until we reach node $C$.

Note that since at the root 0, $\tilde\beta_0(\rho) = P(O(T)|\rho)$ and $\tilde\alpha_0(\rho) = \pi(\rho)$, then a simple way to compute the likelihood for the entire tree is
\beq P(O|\lambda) = \sum_\rho \tilde\beta_0(\rho)\tilde\alpha_0(\rho),\eeq
which again requires $\mathcal O(|T| N^{n+1})$ computations.

\subsection{Solution to problem 2}

Unlike Problem 1, where an exact solution can be given, the solution to Problem 2 is not unique, as it depends on the definition of the ``optimal'' tree of hidden states associated with the observed tree. For example, one optimality criterion is to choose the states $h(C)$ such that individually each state is most likely, which maximizes the expected number of correct hidden states. To solve this problem, we define
\beq \gamma_C (\mu) \equiv P(h(C) = \mu| O,\lambda),\eeq
i.e., the probability of node $C$ being in state $\mu$, given the observed tree $O$ and the model $\lambda$. 

Then the individually most likely state $h(C)$ in terms of $\gamma_C(\mu)$ is
\beq h(C) = \argmax_{\mu}\gamma_C(\mu) .\label{eq:h_optimum_naive}\eeq

However, a problem arises when the HMT has state transitions that are not allowed and the ``optimal'' state tree is not valid, i.e., cannot be generated from such a model. This is due to the fact that Eq.~\eqref{eq:h_optimum_naive} only optimizes for each node $C$ of the tree $T$ and does not explicitly take into account the geometry of $T$ and the structure of the state transitions.

A possible solution to the above problem is to choose a different optimality criterion. For example, one could maximize the expected number of correct states for a nuclear family unit (a node $C$ and its children $\ch(C)$), or simply just the children of node $C$ ($\ch(C)$). Although these criteria might certainly be reasonable depending on the context, the criterion that we propose and expect to be widely applicable is to find the single best state tree $h$, i.e., to maximize $P(h|O,\lambda)$ given the observed tree $O$ and the model $\lambda$, which is equivalent to maximizing $P(h, O|\lambda)$. 

The brute-force method to maximize $P(h|O)$, and hence $P(h, O)$, is to directly compute
\beq P^* \equiv \max_{\mu} P(h = \mu, O) .\eeq
However, this solution requires $\mathcal O(N^{|T|})$ computations, which is expensive and unfeasible. The $\mathcal O(|T| N^{n+1})$ solution that we propose extends the well-known Viterbi algorithm~\cite{viterbi1967error,forney1973viterbi}, as well as the case of independent branches considered in Ref.~\cite{durand2004computational}.

We hence define the best score
\beq \delta_C(\rho) \equiv \max_{\mu(T_D(C))} P(O(T_R(C))), h(T_D(C)) = \mu(T_D(C))| h(C) = \rho) ,\eeq
which is the highest probability, given that node $C$ is in hidden state $\rho$, of observing $O(T_R(C))$, the maximal observed subtree of $T$ rooted at $C$, maximized over the hidden states $h(T_D(C))$ of the descendants of node $C$. In terms of $\delta$, $P^*$ can be written as
\beq P^* = \max_\mu [\delta_0(\mu) \pi_\mu] \label{eq:delta_root}. \eeq
In Appendix~\ref{app:delta}, we show that we can express $\delta_C(\rho)$ recursively for a non-leaf node $C$ as
\beq\delta_C(\rho) = b_\rho(O(C))\max_{\rho_1 \ldots \rho_n} \left[ a^\rho_{\rho_1 \ldots \rho_n} \prod_{i=1}^n\delta_{\ch(C)_i}(\rho_i)\right] , \label{eq:delta_recursion}\eeq
where at a leaf $L$, $\delta_L(\rho) = b_\rho(O(L))$. We compute $\delta_C(\rho)$ for each node by starting from the leaves and working up the tree to the root. Importantly, as we move up the tree for each node, $C$, we store the hidden states of its children, $\rho_1 \ldots \rho_n$, that maximized the value of $\delta_C(\rho)$ in Eq.~\eqref{eq:delta_recursion} for each value of the hidden state of the parent node, $\rho$. At the root, the optimal hidden state, $\mu_m$, is assigned using equation Eq.~\eqref{eq:delta_root}. We then assign the hidden states of the children of the root as the stored hidden states that maximized $\delta_0(\mu_m)$. This process is repeated all the way down the tree to the leaf nodes, assigning the optimal hidden state to each node of the tree. Taken together, this algorithm is a generalization of the Viterbi algorithm to HMT with coupled branches and can be computed efficiently with complexity $\mathcal O(|T| N^{n+1})$.

	\subsection{Solution to Problem 3}
	
	We now turn to Problem 3, the most difficult but perhaps the most practical proposed problem. As in the case of HMMs, given the observed tree as training data, there is no optimal way of estimating the model parameters. However, we propose an iterative procedure, an extension of the Baum-Welch algorithm~\cite{baum1966statistical,baum1967inequality,baum1968growth,baum1970maximization,baum1972inequality} (an example of an expectation-maximization (EM) algorithm~\cite{dempster1977maximum}), that given a tree $T$ of $|T|$ observations and $N$ hidden states, efficiently infers the parameters $\lambda = (a,b,\pi)$. We expect our $\mathcal O(|T| N^{n+1})$ algorithm to locally maximize $P(O|\lambda)$. Moreover, building on the approach used in Refs. \cite{Devijver85} and \cite{durand2004computational}, we propose an algorithm that is numerically stable and does not suffer from the underflow problem.

	We follow an iterative procedure to compute the estimates $\hat\lambda = (\hat a, \hat b, \hat \pi)$, of $\lambda = (a, b, \pi)$:
	
	\begin{enumerate}
		\item Initialize $\lambda$.
		\item Given $\lambda$, compute the estimates $\hat\lambda$.
		\item Set $\lambda = \hat\lambda$.
		\item Repeat Steps 2-3 until convergence.
	\end{enumerate}
	
It is straightforward to use our solutions to Problems 1 and 2 to carry out the expectation maximization iterative procedure. Next, we outline how updated parameters, $\hat a$, $\hat b$, $\hat \pi$, can be estimated from the observed data and the computed $\tilde\alpha$ and $\tilde\beta$.

\subsubsection{Computation of $\hat a$}
	
We estimate $\hat a$ by a variant of simple maximum likelihood estimation:
	\beq \hat a^\rho_{\mu_1\ldots \mu_n} \equiv \frac{\text{expected $\#$ of times parent is in state $\rho$ and $n$ children are in states $\mu_1\ldots\mu_n$}}{\text{expected $\#$ of times parent is in state $\rho$}}.\eeq
	
	To compute $\hat a$, we define the probability $\xi_C(\rho,\mu_1, \ldots, \mu_n)$ as the probability of node $C$ being in state $\rho$, its children $\ch(C)_i$ being in state $\mu_i$, given the observations $O(T)$ and the model $\lambda$, as:
	\beq \xi_C(\rho, \mu_1, \ldots, \mu_n) \equiv P(h(C) = \rho, h(\ch(C)_1) = \mu_1, \ldots, h(\ch(C)_n) = \mu_n | O(T), \lambda) .\eeq
	In terms of $\xi$,
	\beq \hat a^\rho_{\mu_1\ldots\mu_n} = \frac{\sum_\text{nodes $C$} \xi_C(\rho, \mu_1,\ldots \mu_n)}{ \sum_{\nu_1\ldots\nu_n} \sum_\text{nodes $C$}\xi_C(\rho, \nu_1,\ldots,\nu_n)} . \label{eq:a_Hat}\eeq
	It thus suffices to compute $\xi_C(\rho, \mu_1,\ldots,\mu_n)$. By the definition of conditional probability, we can write $\xi_C(\rho, \mu_1,\ldots,\mu_n)$ as
	\beq \xi_C(\rho, \mu_1\ldots\mu_n) = \frac{P(h(C) = \rho, h(\ch(C)_1) = \mu_1, \ldots, h(\ch(C)_n) = \mu_n, O(T) | \lambda)}{P(O(T)| \lambda)} .\eeq
	Since the denominator is simply a normalization factor, we can ignore it. We can write the numerator in terms of $\tilde \alpha$ and $\tilde\beta$ as
	\begin{align}
	&P(h(C) = \rho, h(\ch(C)_1) = \mu_1, \ldots, h(\ch(C)_n) = \mu_n, O(T) | \lambda) = \nonumber \\ &\qquad \tilde\alpha_C(\rho ) a^\rho_{\mu_1\ldots\mu_n} b_{\rho}(O(C))\prod_{i=1}^n b_{\mu_i}(O(\ch(C)_i)) \tilde\beta_{\ch(C)_i}(\mu_i),
	\end{align}
    completing the task at hand.

\subsubsection{Computation of $\hat b$}
		
	We also need a formula for recomputing the observation probability $\hat b_\mu(v)$ given a state $\mu$. We will do this by trying to compute
	\beq \hat b_\mu(v) \equiv \frac{\text{expected $\#$ of times in state $\mu$ and observing $v$}}{\text{expected $\#$ of times in state $\mu$}} .\eeq
	
	In order to compute $\hat b_\mu(v)$, we will need to know the probability of node $C$ being in state $\mu$, which we will call $\gamma_C (\mu)$:
	\beq \gamma_C (\mu) \equiv P(h(C) = \mu| O,\lambda) .\eeq
	In terms of $\gamma$,
	\beq  \hat b_\mu(v) = \frac{\sum_{C \text{ s.t. } O(C) = v} \gamma_C(\mu)}{\sum_C \gamma_C(\mu)} . \label{eq:b_Hat}\eeq
	It thus suffices to compute $\gamma$. By the definition of conditional probability,
	\beq \gamma_C(\mu) = \frac{P(h(C) = \mu, O | \lambda)}{P(O|\lambda)} . \eeq
	Since the denominator is simply a normalization factor, we can rewrite Eq.~\eqref{eq:b_Hat} as
	\beq  \hat b_\mu(v) = \frac{\sum_{C \text{ s.t. } O(C) = v} \tilde\alpha_C(\mu) \tilde\beta_C(\mu)}{\sum_C \tilde\alpha_C(\mu) \tilde\beta_C(\mu)} , \eeq
	where we have used the fact that
	\beq P(h(C) = \mu, O | \lambda) = \tilde\alpha_C(\mu) \tilde\beta_C(\mu).\eeq
	
	\subsubsection{Computation of $\hat\pi$}
	
	We simply estimate $\hat \pi$ as:
	\begin{align}
	\hat\pi(\mu) &\equiv \text{expected $\#$ of times root is in state $\mu$} \nonumber\\
	&= \gamma_0(\mu).
	\end{align}

\subsection{Solution to Problem 3 avoiding the underflow problem}

\subsubsection{Preliminaries}

A practical issue with the solution proposed above is that computing $\alpha$ and $\beta$ requires multiplying together many probabilities. When the number of observations is large, doing so will results in values that eventually exceed the finite machine precision and are rounded to zero (referred to as the underflow problem). To overcome this problem, Devijver~\cite{Devijver85} proposed scaled versions of $\tilde\alpha$ and $\tilde\beta$, called $\alpha$ and $\beta$, respectively, that do not diminish with increasing number of observations. Ref.~\cite{durand2004computational} applied this approach to HMTs, which we now extend to trees with coupled branches. Although we only present the solution to Problem 3 using this new formalism, the results can be easily generalized to Problems 1 and 2. Also, for ease of notation, although all probabilities are assumed to be conditional on the model parameters $\lambda$, we do not show this explicitly.

We begin by defining the following quantities:
\begin{align}
	\beta_C(\rho) &\equiv P(h(C) = \rho | O(T_R(C))) \\
	\alpha_C(\rho) &\equiv \frac{P(O(T_{\overline R}(C))|h(C) = \rho)}{P(O(T_{\overline R}(C))| O(T_R(C)))} .
\end{align}
In what follows, it is useful to note that $P(h(C) = \rho)$ can be defined recursively as
\begin{align}
&P(h(C) = \rho) \nonumber \\
&=\sum_{\mu_0\ldots\mu_{n-1}} \Bigl[(P(h(C) = \rho, h(\s(C))_1 = \mu_1,\ldots,h(\s(C))_{n-1} = \mu_{n-1} | h(\p(C)) = \mu_0) \Bigr.\nonumber \\
&\qquad \qquad \qquad \Bigl.\times P(h(\p(C)) = \mu_0) \Bigr]\nonumber \\
&= \sum_{\mu_0\ldots\mu_{n-1}} a^{\mu_0}_{\rho\mu_1\ldots\mu_{n-1}} P(h(\p(C)) = \mu_0) ,
\label{eq:prob_hidden_state}
\end{align}
where the initialization condition at the root is
\beq P(h(0) = \rho) = \pi(\rho).\eeq

Computing $P(h(C) = \rho)$ for all of the nodes requires $\mathcal O(|T| N^{n+1})$ operations.

\subsubsection{Computation of $\beta$}

$\beta$ is initialized at a leaf $L$ by
\beq \beta_L(\rho) = P(h(L) = \rho | O(L)) = P(O(L) |h(L) = \rho) \frac{P(h(L) = \rho)}{P(O(L))} = b_\rho(O(L)) \frac{P(h(L) = \rho)}{P(O(L))} ,\eeq
where the numerator of the fraction is given in Eq.~\eqref{eq:prob_hidden_state} and its denominator is simply a normalization factor. In Appendix~\ref{app:beta}, we show that for the remaining nodes, $\beta_C(\rho)$ can be expressed recursively as
\beq \beta_C(\rho)  = \frac{b_\rho(O(C)) P(h(C) = \rho) \sum_{\rho_1\ldots\rho_n}\left[\left(\prod_{i=1}^n \frac{\beta_{\ch(C)_i}(\rho_i)}{P(h(\ch(C)_i) = \rho_i)}\right) a^\rho_{\rho_1\ldots\rho_n}\right]}{\sum_\mu b_\mu(O(C)) P(h(C) = \mu) \sum_{\mu_1\ldots\mu_n}\left[\left(\prod_{i=1}^n \frac{\beta_{\ch(C)_i}(\mu_i)}{P(h(\ch(C)_i) = \mu_i)}\right) a^\mu_{\mu_1\ldots\mu_n}\right]} .\eeq

\subsubsection{Computation of $\alpha$}

Here we outline how to compute $\alpha$, where the details are delegated to
Appendix~\ref{app:alpha}. We first show that
\beq \gamma_C(\rho) = \alpha_C(\rho) \beta_C(\rho) . \eeq
We then show that $\gamma_C(\rho)$ can be defined recursively as
\begin{align}
\gamma_C(\rho) &= \frac{ \beta_C(\rho)}{P(h(C) = \rho)} \nonumber \\
&\qquad \times \sum_{\rho_0} \left[\left( \frac{\sum_{\rho_1,\ldots,\rho_{n-1}}a^{\rho_0}_{\rho \rho_1 \ldots \rho_{n-1}} \prod_{i=1}^{n-1} \frac{\beta_{\s(C)_i}(\rho_i)}{P(h(\s(C)_i) = \rho_i)}}{\sum_{\rho'} \frac{\beta_C(\rho')}{P(h(C) = \rho')} \sum_{\rho_1',\ldots,\rho_{n-1}'}  a^{\rho_0}_{\rho' \rho_1' \ldots \rho_{n-1}'} \prod_{i=1}^{n-1} \frac{\beta_{\s(C)_i}(\rho_i')}{P(h(\s(C)_i) = \rho_i')} }\right)\gamma_{\p(C)}(\rho_0)\right] ,
\end{align}
where $\gamma_C(\rho)$ is initialized at the root $0$ of the tree by
\beq \gamma_0(\rho) = P(h(0) = \rho|O(T)) = \beta_0(\rho) .\eeq
Finally, we show that $\alpha_C(\rho)$ can be defined recursively as
\begin{align}
\alpha_C(\rho)) &=  \frac{1}{P(h(C) = \rho)} \nonumber \\
&\times \sum_{\rho_0} \left[\left( \frac{\sum_{\rho_1,\ldots,\rho_{n-1}}a^{\rho_0}_{\rho \rho_1 \ldots \rho_{n-1}} \prod_{i=1}^{n-1} \frac{\beta_{\s(C)_i}(\rho_i)}{P(h(\s(C)_i) = \rho_i)}}{\sum_{\rho'} \frac{\beta_C(\rho')}{P(h(C) = \rho')} \sum_{\rho_1',\ldots,\rho_{n-1}'}  a^{\rho_0}_{\rho' \rho_1' \ldots \rho_{n-1}'} \prod_{i=1}^{n-1} \frac{\beta_{\s(C)_i}(\rho_i')}{P(h(\s(C)_i) = \rho_i')} }\right)\beta_{\p(C)}(\rho_0)\alpha_{\p(C)}(\rho_0)\right] ,
\end{align}
where at the root $\alpha_C(\rho)$ is initialized to be $\alpha_0(\rho) = 1$.

\subsubsection{Computation of $\hat a$}

As before, we define
\beq \hat a^\rho_{\mu_1\ldots\mu_n} = \frac{\sum_\text{nodes $C$} \xi_C(\rho, \mu_1,\ldots \mu_n)}{ \sum_{\nu_1\ldots\nu_n}\sum_\text{nodes $C$}\xi_C(\rho, \nu_1,\ldots,\nu_n)} ,\eeq
where
\begin{align}
	\xi_C(\rho, \mu_1, \ldots, \mu_n) &\equiv P(h(C) = \rho, h(\ch(C)_1) = \mu_1, \ldots, h(\ch(C)_n) = \mu_n, O(T) | \lambda) \nonumber \\
	&= P(O(T_{\overline R}(C)),h(C) = \rho) b_\rho(C)a^\rho_{\mu_1\ldots\mu_n} \nonumber \\
 &\qquad \times \prod_{i=1}^n P(O(T_R(\ch(C)_i))|h(\ch(C)_i) = \mu_i) .
\end{align}
Upon using
\begin{align}
P(O(T_{\overline R}(C)),h(C) = \rho) &=  P(O(T_{\overline R}(C)) | h(C) = \rho) P(h(C) = \rho) \nonumber \\
&= \alpha_C(\rho) P(O(T_{\overline R}(C))| O(T_R(C))) P(h(C) = \rho)
\end{align}
and
\begin{align}
P(O(T_R(\ch(C)_i))|h(\ch(C)_i) = \mu_i) &= P(h(\ch(C)_i) = \mu_i | O(T_R(\ch(C)_i))) \frac{P(O(T_R(\ch(C)_i)))}{P(h(\ch(C)_i) = \mu_i)} \nonumber\\
&= \beta_{\ch(C)_i}(\mu_i) \frac{P(O(T_R(\ch(C)_i)))}{P(h(\ch(C)_i) = \mu_i)} ,
\end{align}
we arrive at
\begin{align}
&\xi_C(\rho, \mu_1, \ldots, \mu_n) = \nonumber \\
&\alpha_C(\rho) P(O(T_{\overline R}(C))| O(T_R(C))) P(h(C) = \rho) b_\rho(C)a^\rho_{\mu_1\ldots\mu_n}\prod_{i=1}^n \beta_{\ch(C)_i}(\mu_i) \frac{P(O(T_R(\ch(C)_i)))}{P(h(\ch(C)_i) = \mu_i)} .
\end{align}
Noting that $P(O(T_{\overline R}(C))| O(T_R(C))) \prod_{i=1}^n P(O(T_R(\ch(C)_i)))$ is simply a normalization factor, we can write
\beq \xi_C(\rho, \mu_1, \ldots, \mu_n) \propto \alpha_C(\rho) P(h(C) = \rho) b_\rho(C)a^\rho_{\mu_1\ldots\mu_n}\prod_{i=1}^n \frac{\beta_{\ch(C)_i}(\mu_i) }{P(h(\ch(C)_i) = \mu_i)}, \eeq
completing the task at hand.

\subsubsection{Computation of $\hat b$}

As described in the previous section,
\beq  \hat b_\mu(v) = \frac{\sum_{C \text{ s.t. } O(C) = v} \gamma_C(\mu)}{\sum_C \gamma_C(\mu)}.\eeq

\subsubsection{Computation of $\hat\pi$}

We again simply estimate $\hat \pi$ as:
\begin{align}
	\hat\pi(\mu) = \gamma_0(\mu).
\end{align}

\section{Simulations}
 
   	In this section we test the algorithms presented in Secs.~\ref{sec:algorithms} on simulated trees. We first check the validity of the algorithms, and then show how self-consistency checks can be used to ensure that the assumptions used for the inference are correct. All code used for our analysis is available on the Hormoz Lab GitLab page (\href{https://gitlab.com/hormozlab/hmt}{https://gitlab.com/hormozlab/hmt}).

\begin{figure}[htbp]
     \centering
     \includegraphics[width=\linewidth]{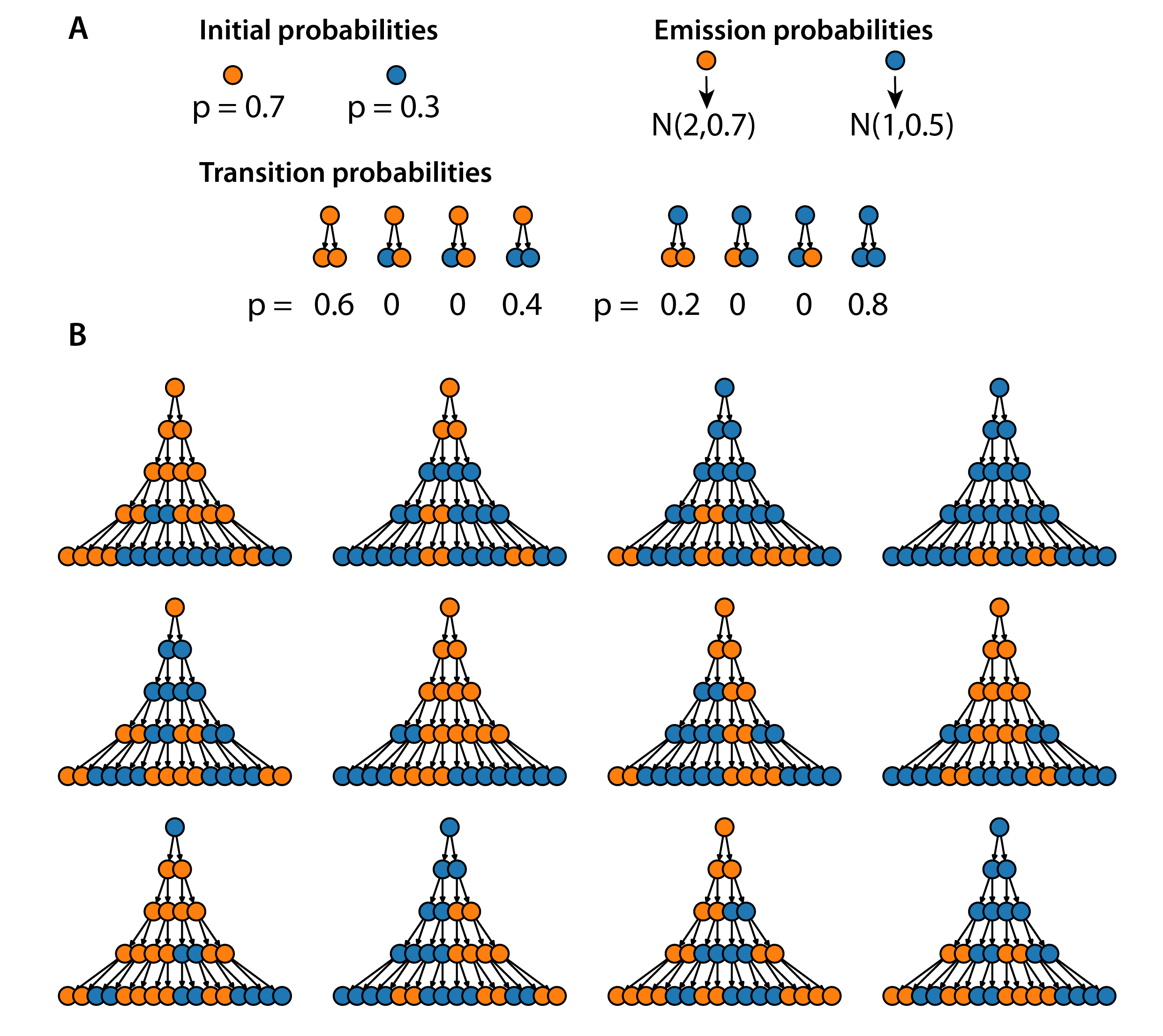}
     \caption{A. The model parameters used to generate the simulated trees. There are two hidden states shown as orange and blue circles. The state of the root node of each tree is assigned to one of the hidden states with the shown probabilities. The observed value of each node is drawn from a Gaussian distribution with the mean and standard deviation determined by the hidden state of the node. The hidden states of the children are assigned probabilistically conditional on the state of the parent node with the transition probabilities shown. We chose the transition probabilities such that the sibling nodes are always in the same hidden state (are perfectly coupled). B. Examples of simulated trees with the hidden state of each node visualized.}
     \label{fig:trees}
 \end{figure}

    We first generated simulated binary trees where each node can be in one of two possible hidden states. Conditional on the hidden state, the observable on each node is scalar drawn from a Gaussian distribution. The probability that the root of the tree is in a given state and the probability of the hidden state of the children conditional on the hidden state of their parent is shown in Fig.~\ref{fig:trees}A. Importantly, the transition probabilities of the hidden state of a parent node to the children is chosen so that the states of the children are coupled. In our example, the states of two sibling nodes are always identical. We simulated 150 trees of 5 generations (32 nodes for each tree).

\begin{figure}[htbp]
     \centering
     \includegraphics[width=\linewidth]{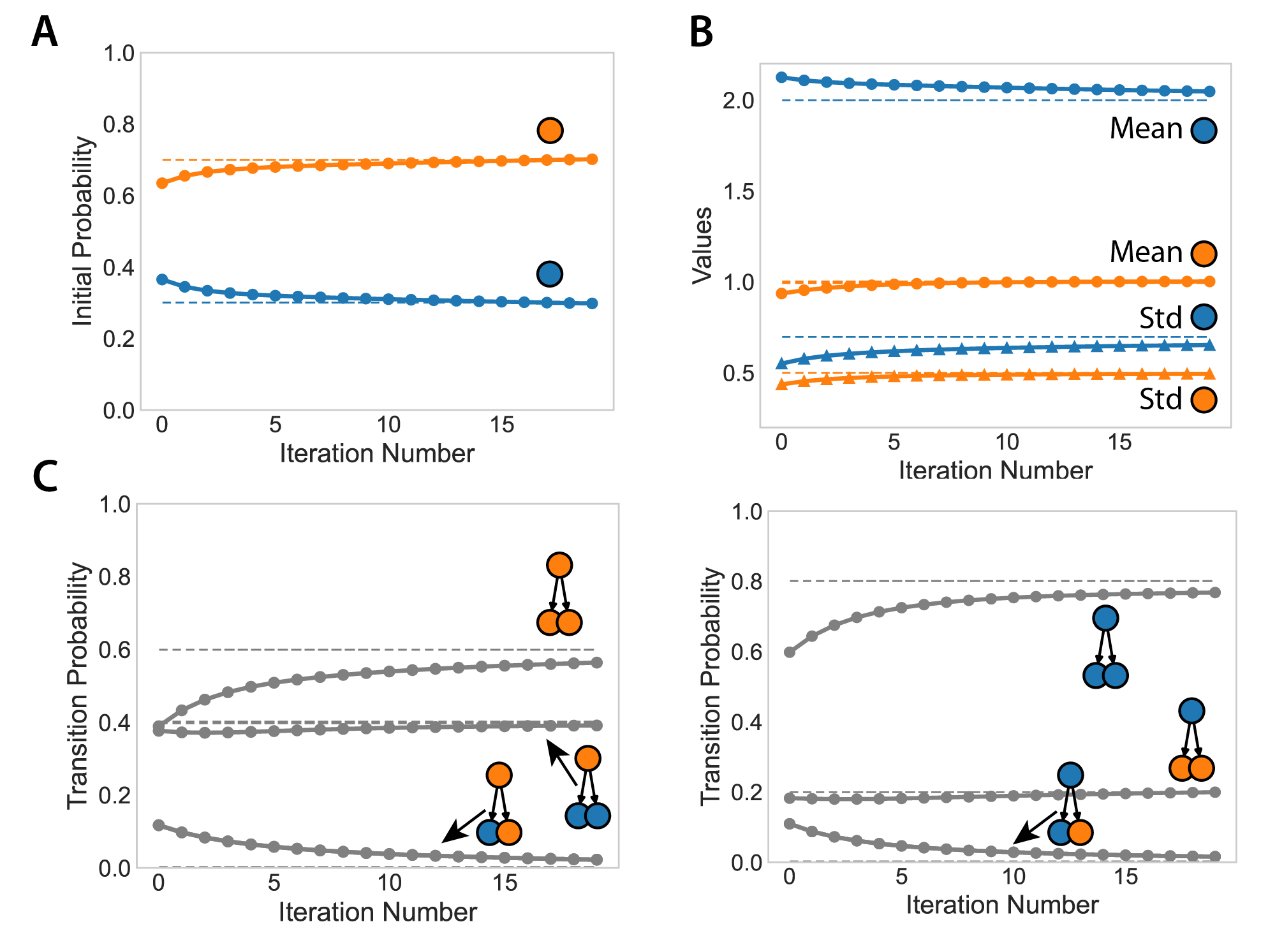}
     \caption{A. Learned initial probabilities of the hidden state of the root of the trees after 20 iterations of our expectation maximization algorithm applied to 150 simulated trees of 5 generations. The learned shedding probabilities and transition probabilities as a function of the number of iterations is shown in panels B and C respectively. In all panels, the dashed lines show the true parameter value.}
     \label{fig:trees_inference}
 \end{figure}

    Next, we used the observed values of the simulated trees to learn the model parameters (initial probabilities, emission probabilities, and transition probabilities) using the solution to Problem 3 outlined in the previous section. The initial set of parameters was estimated by aggregating all the observed values across the nodes and applying k-mean clustering to them to assign each observed value to one of two states. The assignments of the nodes then was used to estimate the probability that the root of the trees were in a given hidden state (initial probabilities), the mean and standard deviation of the Gaussian of the observed value conditional on each hidden state (emission probabilities), and the probabilities of the hidden states of the children conditional on the state of their parent (transition probabilities). Fig.~\ref{fig:trees_inference} shows the learned parameters using our expectation-maximization algorithm as a function of the number of iterations of the algorithm. As shown, the algorithm correctly learns the true parameter values. Importantly, this computation is done efficiently, requiring only minutes on a single CPU.

    A fundamental limitation of inference problems is that the inference always learns the model parameters even if the assumptions going into the inference model are wrong. Ideally, we would like to know if our model assumptions are not consistent with the observed data. Some of the key assumptions made in HMT models are the number of hidden states, their Markovian nature, and that the transitions rates remain constant over time. We can check that our model assumptions are consistent with the observed data by learning the parameters of the model form the data, generating simulated data using the learned parameters, and then comparing summary statistics of the simulated data to the actual data.

    \begin{figure}[htbp]
     \centering
     \includegraphics[width=\linewidth]{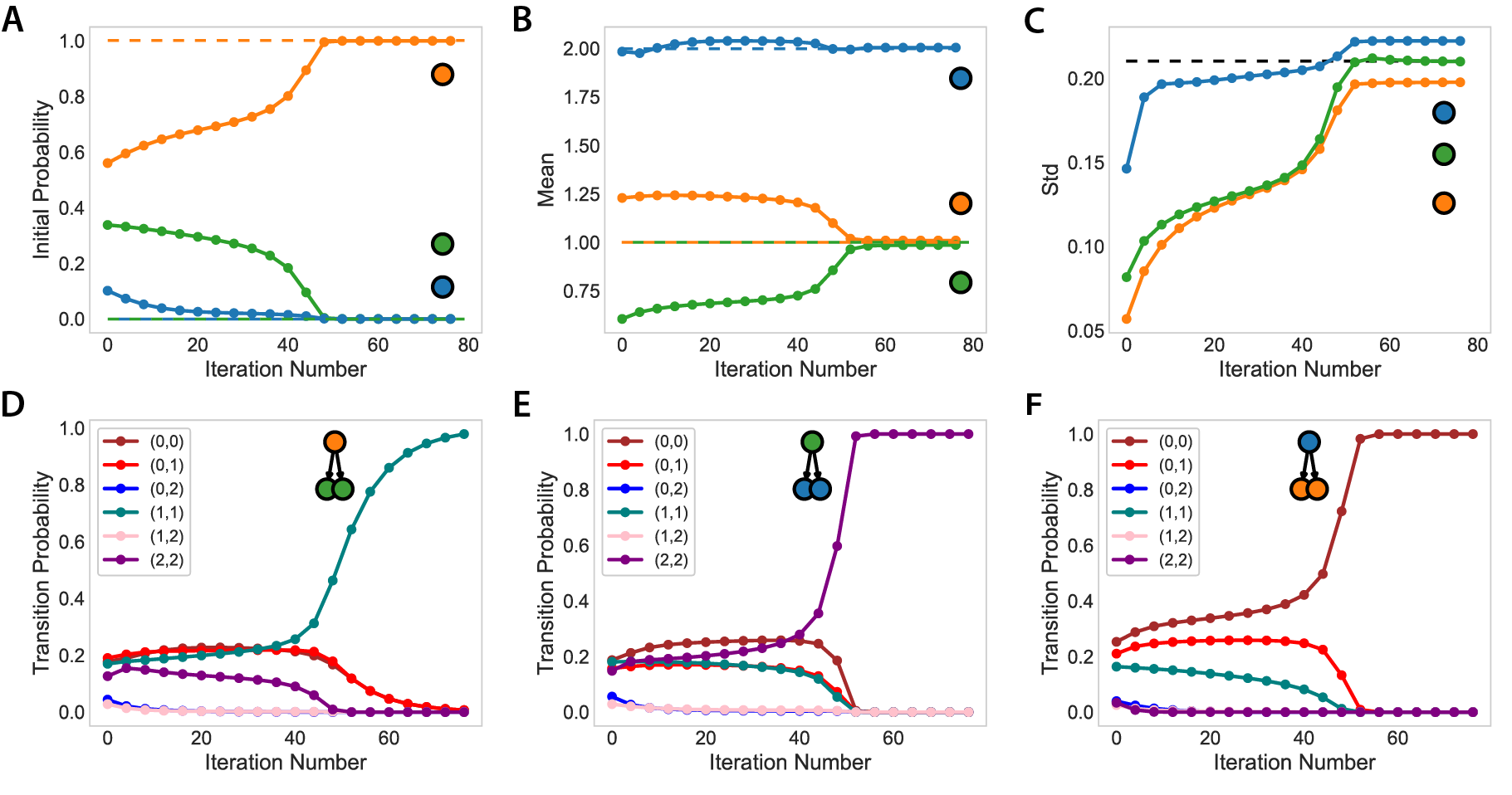}
     \caption{A. Learned initial probabilities of the hidden state of the root of the trees after 80 iterations of our expectation maximization algorithm with three hidden states applied to 150 simulated trees of 5 generations. The three hidden states 0, 1, and 2 are shown as orange, green, and blue respectively. The learned shedding probabilities as a function of the number of iterations is shown in panels B and C, and the learned transition probabilities as a function of the number of iterations is shown in panels D-F. In all panels, the dashed lines show the true parameter value. Panels D, E, and F show the inferred transition rates from a parent node in states 0, 1, and 2 respectively to all possible combinations of states for the children. All transitions rates vanish except for the one allowed transition.}
     \label{fig:trees_3_inference}
\end{figure}

\begin{figure}[htbp]
     \centering
     \includegraphics[width=\linewidth]{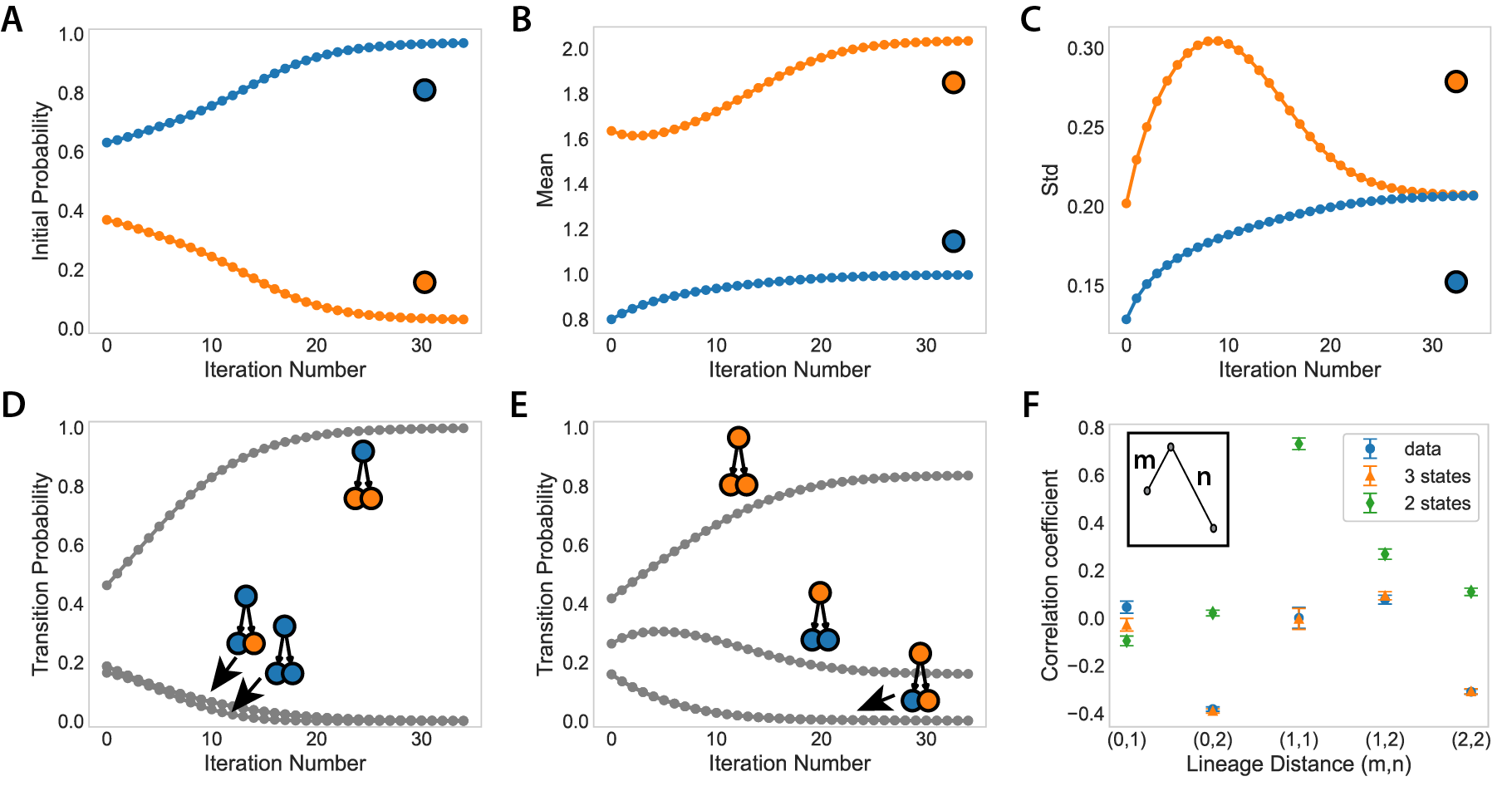}
     \caption{A. Learned initial probabilities of the hidden state of the root of the trees after 35 iterations of our expectation maximization algorithm with two hidden states applied to 150 simulated trees of 5 generations. The learned shedding probabilities as a function of the number of iterations is shown in panels B and C, and the learned transition probabilities as a function of the number of iterations is shown in panels D and E. F. Pair-wise linear Pearson correlation coefficients for true model vs learned two and three state models. Inset: lineage distance $(m,n)$ for two nodes denotes the distance $m$ and $n$ from each node to the two node's most recent common ancestor. For example, sibling nodes are denoted as (1,1) and parent-child nodes as (0,1). The correlations predicted by the 3-state inferred model are consistent with that of the data whereas those from the 2-state model are not.}
     \label{fig:trees_2_inference}
\end{figure}
    
    Here, we demonstrate this approach by generating trees with a true model that has three hidden states, but only assume two hidden states during the inference. To make this task more difficult, we also assumed that two of the hidden states share the same emission probabilities and are therefore indistinguishable based on observing a single node. In particular, we use three hidden states 0, 1, and 2. Transitions are deterministic and the state of the child notes are perfectly correlated. If a parent is in state 0, both children will be in state 1. If a parent in state 1, both children will be in state 2. And finally, if a parent is in state 2, both children will be in state 0. All emission probabilities are assumed to be normal distributions parameterized by a mean and standard deviation. We simulated 150 trees of 5 generations (32 nodes for each tree). Fig.~\ref{fig:trees_3_inference} shows the learned parameters using our expectation-maximization algorithm with the model assuming the correct number of three hidden states. As shown, the algorithm correctly learns the true parameter values. Fig.~\ref{fig:trees_2_inference} shows the learned parameters using our expectation-maximization algorithm using an inference model that assumes that there are only two hidden states. The algorithm still converges to some parameter values. However, it is possible to show that the two-state model cannot describe the observed data by checking self-consistency. To do so, we simulated trees with the parameters of the learned two-state model and computed the Pearson correlation between the observed values on two nodes of the trees as a function of their lineage distance. The correlations computed using trees simulated with the inferred parameters of the two-state model are not consistent with the observed correlations in the data. Therefore, the assumptions of the inference model, the number of hidden states in this case, were not correct. In summary, self-consistency checks can be used to check the validity of assumptions used in the inference model.

\section{Summary}

In this paper, we introduced an algorithm for solving hidden Markov models on trees with coupled branches, a scenario commonly encountered in biological systems where interdependencies between related entities (e.g., cells, genetic loci) cannot be ignored. Our approach extends traditional algorithms that are limited to tree structures with independent branches, therefore providing a more realistic modeling of hierarchical biological processes.

Importantly, the complexity of solving HMMs on trees does not necessarily become intractable when branches are coupled. Specifically, we found that the computational cost increases only polynomially from $\mathcal{O}(TN^2)$ to $\mathcal{O}(TN^3)$ for binary trees, where $T$ is the number of the nodes in the tree and $N$ the number of hidden states. This is a significant finding as it suggests that even with the added complexity of coupled branches, the problem remains computationally feasible for a reasonable number of states and tree sizes.

Our method is useful for the modeling of biological systems. For example, in cellular lineage studies, cells derived from the same progenitor often exhibit dependencies in their phenotypic traits due to shared cytoplasmic contents or genetic material. Traditional independent branch models fail to capture such dependencies, which can lead to incorrect inferences about cellular dynamics. By incorporating branch coupling into the tree HMM framework, our approach allows for a more accurate representation of the underlying biological processes, potentially leading to better predictions of cellular behavior.

\acknowledgments{It is a pleasure to acknowledge helpful conversations with Keyon Vafa. This work is partially supported by the Center for Mathematical Sciences and Applications at Harvard University (F. V.). This work was also supported by funding from the National Institutes of Health (NIH) National Heart, Lung, and Blood Institute grant no. R01HL158269 and R01HL158192.}

\bibliography{refs}

\appendix

\section{Derivations}
\label{app:derivations}

\subsection{Computation of $\delta$}
\label{app:delta}

Here we show that $\delta_C(\rho)$ can be defined recursively as
\beq \delta_C(\rho)= b_\rho(C)\max_{\rho_1 \ldots \rho_n} \left[ a^\rho_{\rho_1 \ldots \rho_n} \prod_{i=1}^n\delta_{\ch(C)_i}(\rho_i)\right] .\eeq
We compute:
\begin{align}
	&\delta_C(\rho) \nonumber \\
    &=\max_{\mu(T_D(C))} P(O(T_R(C)), h(T_D(C) = \mu(T_D(C))| h(C) = \rho) \nonumber \\
	&= \max_{\mu(\ch(C))}\max_{\mu(T_D(\ch(C)))} \nonumber \\
	& \qquad P(O(C),O(T_D(C)), h(\ch(C)) = \mu(\ch(C)), h(T_D(\ch(C)) = \mu(T_D(\ch(C)))| h(C) = \rho)) \nonumber \\
	&= P(O(C)| h(C) = \rho)\max_{\mu(\ch(C))}\max_{\mu(T_D(\ch(C)))} \nonumber \\
	& \qquad P(O(T_D(C)), h(\ch(C)) = \mu(\ch(C)), h(T_D(\ch(C)) = \mu(T_D(\ch(C)))| h(C) = \rho) \nonumber \\
	&= b_\rho(C)\max_{\mu(\ch(C))} \Biggl[ P(h(\ch(C)) = \mu(\ch(C))|h(C) = \rho) \max_{\mu(T_D(\ch(C)))}\Biggr.\nonumber \\
	& \qquad \quad \Biggl. P(O(T_D(C)), h(T_D(\ch(C)) = \mu(T_D(\ch(C)))| h(\ch(C)) = \mu(\ch(C)))\Biggr] \nonumber \\
	&= b_\rho(C)\max_{\rho_1 \ldots \rho_n} \Biggl[ a^\rho_{\rho_1 \ldots \rho_n} \max_{\mu(T_D(\ch(C)))} \Biggr.\nonumber \\
	& \qquad \Biggl.  P(O(T_D(C)), h(T_D(\ch(C)) = \mu(T_D(\ch(C)))| h(\ch(C)_1) = \rho_1,\ldots,h(\ch(C)_n) = \rho_n)\Biggr] \nonumber \\
	&= b_\rho(C)\max_{\rho_1 \ldots \rho_n} \Biggl[ a^\rho_{\rho_1 \ldots \rho_n} \max_{\mu(T_D(\ch(C)))} \Biggr.\nonumber \\
	& \qquad \Biggl. 
	\prod_{i=1}^n P(O(\ch(C)_i), O(T_D(\ch(C)_i)), h(T_D(\ch(C)_i)) = \mu(T_D(\ch(C)_i))| h(\ch(C)_i) = \rho_i)\Biggr] \nonumber \\
	&= b_\rho(C)\max_{\rho_1 \ldots \rho_n} \left[ a^\rho_{\rho_1 \ldots \rho_n} \prod_{i=1}^n\delta_{\ch(C)_i}(\rho_i)\right] ,
\end{align}
completing the task at hand.

\subsection{Computation of $\beta$}
\label{app:beta}

Here we show that for non-leaves, $\beta_C(\rho)$ can be expressed recursively as
\beq  \beta_C(\rho)  = \frac{b_\rho(O(C)) P(h(C) = \rho) \sum_{\rho_1\ldots\rho_n}\left[\left(\prod_{i=1}^n \frac{\beta_{\ch(C)_i}(\rho_i)}{P(h(\ch(C)_i) = \rho_i)}\right) a^\rho_{\rho_1\ldots\rho_n}\right]}{\sum_\mu b_\mu(O(C)) P(h(C) = \mu) \sum_{\mu_1\ldots\mu_n}\left[\left(\prod_{i=1}^n \frac{\beta_{\ch(C)_i}(\mu_i)}{P(h(\ch(C)_i) = \mu_i)}\right) a^\mu_{\mu_1\ldots\mu_n}\right]} .\eeq
We have:
\begin{align}
\beta_C(\rho) &= P(h(C) = \rho | O(T_R(C))) \nonumber \\
&= P(O(T_R(C)) |h(C) = \rho) \frac{P(h(C) = \rho)}{P(O(T_R(C)))} \nonumber \\
&= P(O(C)|h(C) = \rho) P(O(T_R(\ch(C)_1)),\ldots,O(T_R(\ch(C)_n)) | h(C) = \rho) \frac{P(h(C) = \rho)}{P(O(T_R(C)))} \nonumber \\
&= b_\rho(O(C)) \sum_{\rho_1\ldots\rho_n}\left[\left(\prod_i^n P(O(T_R(\ch(C)_i)) | h(\ch(C)_i) = \rho_i)\right) a^\rho_{\rho_1\ldots\rho_n}\right]\frac{P(h(C) = \rho)}{P(O(T_R(C)))} \nonumber \\
&= b_\rho(O(C)) \frac{P(h(C) = \rho)}{P(O(T_R(C)))} \nonumber \\
&\qquad \times \sum_{\rho_1\ldots\rho_n}\left[\left(\prod_i^n P(h(\ch(C)_i) = \rho_i |O(T_R(\ch(C)_i))) \frac{P(O(T_R(\ch(C)_i)))}{P(h(\ch(C)_i) = \rho_i)}\right) a^\rho_{\rho_1\ldots\rho_n}\right] \nonumber \\
&= b_\rho(O(C)) \sum_{\rho_1\ldots\rho_n}\left[\left(\prod_i^n \beta_{\ch(C)_i}(\rho_i) \frac{P(O(T_R(\ch(C)_i)))}{P(h(\ch(C)_i) = \rho_i)}\right) a^\rho_{\rho_1\ldots\rho_n}\right]\frac{P(h(C) = \rho)}{P(O(T_R(C)))} \nonumber \\
&= b_\rho(O(C)) P(h(C) = \rho) \sum_{\rho_1\ldots\rho_n}\left[\left(\prod_{i=1}^n \frac{\beta_{\ch(C)_i}(\rho_i)}{P(h(\ch(C)_i) = \rho_i)}\right) a^\rho_{\rho_1\ldots\rho_n}\right] \nonumber \\
&\qquad \times\frac{\prod_{i=1}^nP(O(T_R(\ch(C)_i)))}{P(O(T_R(C)))} \nonumber \\
&= \frac{b_\rho(O(C)) P(h(C) = \rho) \sum_{\rho_1\ldots\rho_n}\left[\left(\prod_{i=1}^n \frac{\beta_{\ch(C)_i}(\rho_i)}{P(h(\ch(C)_i) = \rho_i)}\right) a^\rho_{\rho_1\ldots\rho_n}\right]}{\sum_\mu b_\mu(O(C)) P(h(C) = \mu) \sum_{\mu_1\ldots\mu_n}\left[\left(\prod_{i=1}^n \frac{\beta_{\ch(C)_i}(\mu_i)}{P(h(\ch(C)_i) = \mu_i)}\right) a^\mu_{\mu_1\ldots\mu_n}\right]} .
\end{align}
In the last step we used the normalization condition $\sum_\rho\beta_C(\rho) = 1$.

\subsection{Computation of $\alpha$}
\label{app:alpha}

Before we compute $\alpha_C(\rho)$ , we will first show that
\beq \gamma_C(\rho) =  P(h(C) = \rho | O(T)) =  \frac{P(O(T_{\overline R}(C))|h(C) = \rho)}{P(O(T_{\overline R}(C))| O(T_R(C)))}  P(h(C) = \rho | O(T_R(C))) = \alpha_C(\rho) \beta_C(\rho) .\eeq
We will then show that $\gamma_C(\rho)$ can be defined recursively as
\begin{align}
\gamma_C(\rho) &= \frac{ \beta_C(\rho)}{P(h(C) = \rho)} \nonumber \\
&\qquad \times \sum_{\rho_0} \left[\left( \frac{\sum_{\rho_1,\ldots,\rho_{n-1}}a^{\rho_0}_{\rho \rho_1 \ldots \rho_{n-1}} \prod_{i=1}^{n-1} \frac{\beta_{\s(C)_i}(\rho_i)}{P(h(\s(C)_i) = \rho_i)}}{\sum_{\rho'} \frac{\beta_C(\rho')}{P(h(C) = \rho')} \sum_{\rho_1',\ldots,\rho_{n-1}'}  a^{\rho_0}_{\rho' \rho_1' \ldots \rho_{n-1}'} \prod_{i=1}^{n-1} \frac{\beta_{\s(C)_i}(\rho_i')}{P(h(\s(C)_i) = \rho_i')} }\right)\gamma_{\p(C)}(\rho_0)\right] ,
\end{align}
where $\gamma_C(\rho)$ is initialized at the root of the tree $0$ by
\beq \gamma_0(\rho) = P(h(0) = \rho|O(T)) = \beta_0(\rho) .\eeq
Finally, we will have then shown that $\alpha_C(\rho)$ can be defined recursively as
\begin{align}
\alpha_C(\rho)) &= \frac{1}{P(h(C) = \rho)} \nonumber \\
&\times\sum_{\rho_0} \left[\left( \frac{\sum_{\rho_1,\ldots,\rho_{n-1}}a^{\rho_0}_{\rho \rho_1 \ldots \rho_{n-1}} \prod_{i=1}^{n-1} \frac{\beta_{\s(C)_i}(\rho_i)}{P(h(\s(C)_i) = \rho_i)}}{\sum_{\rho'} \frac{\beta_C(\rho')}{P(h(C) = \rho')} \sum_{\rho_1',\ldots,\rho_{n-1}'}  a^{\rho_0}_{\rho' \rho_1' \ldots \rho_{n-1}'} \prod_{i=1}^{n-1} \frac{\beta_{\s(C)_i}(\rho_i')}{P(h(\s(C)_i) = \rho_i')} }\right)\beta_{\p(C)}(\rho_0)\alpha_{\p(C)}(\rho_0)\right] ,
\end{align}
where at the root $\alpha_C(\rho)$ is initialized to be $\alpha_0(\rho) = 1$.

We first note that since $O(T)$ can be decomposed as $O(T) = \left\{T_R(C)),O(T_{\overline R}(C)\right\}$, then by the definition of conditional probability,
\beq \gamma_C(\rho) = P(h(C) = \rho | O(T_R(C)),O(T_{\overline R}(C)) = \frac{P(h(C) = \rho, O(T_{\overline R}(C))|O(T_R(C)))}{P(O(T_{\overline R}(C)|O(T_R(C)))} . \eeq
By Bayes' rule,
\begin{align} 
&P(h(C) = \rho, O(T_{\overline R}(C))|O(T_R(C))) \nonumber\\
& \qquad = P(O(T_R(C)) |h(C) = \rho, O(T_{\overline R}(C)))\frac{P(h(C) = \rho, O(T_{\overline R}(C)))}{P(O(T_R(C)))} .
\end{align}
By the Markov property,
\beq P(O(T_R(C)) |h(C) = \rho, O(T_{\overline R}(C))) = P(O(T_R(C)) |h(C) = \rho), \eeq
and thus
\begin{align}
&P(h(C) = \rho, O(T_{\overline R}(C))|O(T_R(C))) \nonumber \\
&\qquad = P(O(T_R(C)) |h(C) = \rho) \frac{P(h(C) = \rho, O(T_{\overline R}(C)))}{P(O(T_R(C)))} \nonumber \\
&\qquad = \frac{P(O(T_R(C)) | h(C) = \rho)}{P(O(T_R(C)))} P(h(C) = \rho| O(T_{\overline R}(C))) P(O(T_{\overline R}(C))) \nonumber \\
&\qquad = \frac{P(O(T_R(C)) |h(C) = \rho) P(h(C) = \rho)}{P(O(T_R(C)))} \frac{P(h(C) = \rho| O(T_{\overline R}(C))) P(O(T_{\overline R}(C)))}{P(h(C) = \rho)} \nonumber \\
&\qquad = P(h(C) = \rho | O(T_R(C))) P(O(T_{\overline R}(C))|h(C) = \rho) \nonumber \\
&\qquad =  \alpha_C(\rho) \beta_C(\rho) P(O(T_{\overline R}(C)|O(T_R(C))) .
\end{align}
Therefore,
\beq \gamma_C(\rho) = \frac{P(h(C) = \rho, O(T_{\overline R}(C))|O(T_R(C)))}{P(O(T_{\overline R}(C)|O(T_R(C)))} = \alpha_C(\rho) \beta_C(\rho) ,\eeq
completing the task at hand. $\gamma_C(\rho)$ is initialized at the root of the tree $0$ by
\beq \gamma_0(\rho) = P(h(0) = \rho|O(T)) = \beta_0(\rho).\eeq
For each of the remaining nodes, $\gamma_C(\rho)$ can be expressed recursively as
\begin{align}
	&\gamma_C(\rho) = P(h(C) = \rho | O(T)) \nonumber \\
	&= \sum_{\rho_0} P(h(C) = \rho, h(\p(C) = \rho_0 | O(T))  \nonumber\\
	&= \sum_{\rho_0} \frac{P(h(C) = \rho, h(\p(C) = \rho_0 , O(T))}{P(O(T))}  \nonumber\\
	&=\sum_{\rho_0} \frac{P(h(C) = \rho, h(\p(C)) = \rho_0,  O(T))}{P(h(\p(C)) = \rho_0, O(T))}P(h(\p(C)) = \rho_0 |O(T)) \nonumber \\
	&=\sum_{\rho_0} \frac{P(h(C) = \rho, h(\p(C)) = \rho_0,  O(T))}{\sum_{\rho'} P(h(C) = \rho', h(\p(C)) = \rho_0, O(T))}\gamma_{\p(C)}(\rho_0) \nonumber \\
	&=\sum_{\rho_0} \Biggl[ \gamma_{\p(C)}(\rho_0) \times \Biggr.\nonumber \\ 
 &\quad \frac{\sum_{\rho_1,\ldots,\rho_{n-1}}P(h(C) = \rho, h(\p(C)) = \rho_0, h(\s(C)_1) = \rho_1, \ldots, h(\s(C)_{n-1}) = \rho_{n-1}, O(T))}{\sum_{\rho'} \sum_{\rho_1',\ldots,\rho_{n-1}'} P(h(C) = \rho', h(\p(C)) = \rho_0, h(\s(C)_1) = \rho_1', \ldots, h(\s(C)_{n-1}) = \rho_{n-1}', O(T))}  \Biggr]
	\label{eq:hard1}
\end{align}

It thus suffices to compute $P(h(C) = \rho, h(\p(C)) = \rho_0, h(\s(C)_1) = \rho_1, \ldots, h(\s(C)_{n-1}) = \rho_{n-1}, O(T))$, which we do now. We first show that
\begin{align}
&P(h(C) = \rho, h(\p(C)) = \rho_0, h(\s(C)_1) = \rho_1, \ldots, h(\s(C)_{n-1}) = \rho_{n-1}, O(T)) \nonumber \\
&\qquad = P(O(T) | h(C) = \rho, h(\p(C)) = \rho_0, h(\s(C)_1) = \rho_1, \ldots, h(\s(C)_{n-1}) = \rho_{n-1}) \nonumber \\
&\qquad\qquad \times a^{\rho_0}_{\rho \rho_1 \ldots \rho_{n-1}}P( h(\p(C)) = \rho_0) ,
\end{align}
which we do so now. By the chain rule of probability,
\begin{align}
&P(h(C) = \rho, h(\p(C)) = \rho_0, h(\s(C)_1) = \rho_1, \ldots, h(\s(C)_{n-1}) = \rho_{n-1}, O(T)) \nonumber \\
&\qquad =P(O(T) | h(C) = \rho, h(\p(C)) = \rho_0, h(\s(C)_1) = \rho_1, \ldots, h(\s(C)_{n-1}) = \rho_{n-1}) \nonumber \\
&\qquad \qquad \times P( h(C) = \rho, h(\p(C)) = \rho_0, h(\s(C)_1) = \rho_1, \ldots, h(\s(C)_{n-1}) = \rho_{n-1}) .
\end{align}
By the Markov property, 
\begin{align}
&P( h(C) = \rho, h(\p(C)) = \rho_0, h(\s(C)_1) = \rho_1, \ldots, h(\s(C)_{n-1}) = \rho_{n-1}) \nonumber \\
&=  P( h(C) = \rho, h(\s(C)_1) = \rho_1, \ldots, h(\s(C)_{n-1}) = \rho_{n-1} |  h(\p(C)) = \rho_0) P( h(\p(C)) = \rho_0) \nonumber \\
&= a^{\rho_0}_{\rho \rho_1 \ldots \rho_{n-1}}P( h(\p(C)) = \rho_0) .
\end{align}
Hence
\begin{align}
&P(h(C) = \rho, h(\p(C)) = \rho_0, h(\s(C)_1) = \rho_1, \ldots, h(\s(C)_{n-1}) = \rho_{n-1}, O(T)) \nonumber \\
&\qquad= P(O(T) | h(C) = \rho, h(\p(C)) = \rho_0, h(\s(C)_1) = \rho_1, \ldots, h(\s(C)_{n-1}) = \rho_{n-1})  \nonumber \\
&\qquad\qquad \times a^{\rho_0}_{\rho \rho_1 \ldots \rho_{n-1}}P( h(\p(C)) = \rho_0) .
\end{align}

We now focus on the first term. Noting that we can decompose $O(T)$ as
\beq O(T) = \{O(T_{\overline D}(\p(C))), O(T_R(C)), O(T_R(\s(C)_1)), \ldots, O(T_R(\s(C)_{n-1})) \} ,\eeq
where $T_{\overline D}(\p(C))$ is the maximal subtree of $T$ with $\p(C)$ as a leaf,
by the Markov property, we can write
\begin{align}
&P(O(T) | h(C) = \rho, h(\p(C)) = \rho_0, h(\s(C)_1) = \rho_1, \ldots, h(\s(C)_{n-1}) = \rho_{n-1}) \nonumber \\
&= P(O(T_{\overline D}(\p(C))) |  h(\p(C)) = \rho_0) P(O(T_R(C)) | h(C) = \rho)
\prod_{i=1}^{n-1} P(O(T_R(\s(C)_i)) | h(\s(C)_i) = \rho_i) .
\end{align}
By Bayes' rule, it follows that the product of the second and third terms in the above equation can be written as
\begin{align}
&P(O(T_R(C)) | h(C) = \rho)\prod_{i=1}^{n-1} P(O(T_R(\s(C)_i)) | h(\s(C)_i) = \rho_i)  =\nonumber \\
&P(h(C) = \rho | O(T_R(C))) \frac{P( O(T_R(C)))}{P(h(C) = \rho)} \prod_{i=1}^{n-1} P(h(\s(C)_i) = \rho_i| O(T_R(\s(C)_i))) \frac{P( O(T_R(\s(C)_i)))}{P(h(\s(C)_i) = \rho_i)} .
\end{align}

In terms of $\beta$,
\begin{align}
&P(O(T) | h(C) = \rho, h(\p(C)) = \rho_0, h(\s(C)_1) = \rho_1, \ldots, h(\s(C)_{n-1}) = \rho_{n-1}) \nonumber \\
& = \beta_C(\rho) P(O(T_{\overline D}(\p(C))) | h(\p(C)) = \rho_0)  \frac{P( O(T_R(C)))}{P(h(C) = \rho)} \prod_{i=1}^{n-1} \beta_{\s(C)_i}(\rho_i)\frac{P( O(T_R(\s(C)_i)))}{P(h(\s(C)_i) = \rho_i)} .
\end{align}
Now putting everything together, we have
\begin{align}
\gamma_C(\rho) &= \sum_{\rho_0} \left[\left( \frac{\sum_{\rho_1,\ldots,\rho_{n-1}}\frac{ \beta_C(\rho)}{P(h(C) = \rho)} a^{\rho_0}_{\rho \rho_1 \ldots \rho_{n-1}} \prod_{i=1}^{n-1} \frac{\beta_{\s(C)_i}(\rho_i)}{P(h(\s(C)_i) = \rho_i)}}{\sum_{\rho'} \sum_{\rho_1',\ldots,\rho_{n-1}'} \frac{\beta_C(\rho')}{P(h(C) = \rho')} a^{\rho_0}_{\rho' \rho_1' \ldots \rho_{n-1}'} \prod_{i=1}^{n-1} \frac{\beta_{\s(C)_i}(\rho_i')}{P(h(\s(C)_i) = \rho_i')} }\right)\gamma_{\p(C)}(\rho_0)\right]\nonumber \\
&= \frac{ \beta_C(\rho)}{P(h(C) = \rho)} \nonumber \\
&\qquad \times \sum_{\rho_0} \left[\left( \frac{\sum_{\rho_1,\ldots,\rho_{n-1}}a^{\rho_0}_{\rho \rho_1 \ldots \rho_{n-1}} \prod_{i=1}^{n-1} \frac{\beta_{\s(C)_i}(\rho_i)}{P(h(\s(C)_i) = \rho_i)}}{\sum_{\rho'} \frac{\beta_C(\rho')}{P(h(C) = \rho')} \sum_{\rho_1',\ldots,\rho_{n-1}'}  a^{\rho_0}_{\rho' \rho_1' \ldots \rho_{n-1}'} \prod_{i=1}^{n-1} \frac{\beta_{\s(C)_i}(\rho_i')}{P(h(\s(C)_i) = \rho_i')} }\right)\gamma_{\p(C)}(\rho_0)\right] .
\end{align}
We find that $\alpha_C(\rho)$ is expressed recursively as
\begin{align}
\alpha_C(\rho)) &= \frac{\gamma_C(\rho)}{\beta_C(\rho)} \nonumber \\
&= \frac{1}{P(h(C) = \rho)} \nonumber \\
&\quad \times \sum_{\rho_0} \left[\left( \frac{\sum_{\rho_1,\ldots,\rho_{n-1}}a^{\rho_0}_{\rho \rho_1 \ldots \rho_{n-1}} \prod_{i=1}^{n-1} \frac{\beta_{\s(C)_i}(\rho_i)}{P(h(\s(C)_i) = \rho_i)}}{\sum_{\rho'} \frac{\beta_C(\rho')}{P(h(C) = \rho')} \sum_{\rho_1',\ldots,\rho_{n-1}'}  a^{\rho_0}_{\rho' \rho_1' \ldots \rho_{n-1}'} \prod_{i=1}^{n-1} \frac{\beta_{\s(C)_i}(\rho_i')}{P(h(\s(C)_i) = \rho_i')} }\right)\beta_{\p(C)}(\rho_0)\alpha_{\p(C)}(\rho_0)\right] .
\end{align}

\end{document}